\documentclass[conference]{IEEEtran}
\IEEEoverridecommandlockouts
\usepackage{cite}
\usepackage{amsmath,amssymb,amsfonts}
\usepackage{graphicx}
\usepackage{textcomp}
\usepackage{multirow}
\usepackage{xcolor}
\usepackage{CJKutf8}
\usepackage{hyperref}
\usepackage{amsmath}
\usepackage{algorithm}
\usepackage{algpseudocode}

\def\BibTeX{{\rm B\kern-.05em{\sc i\kern-.025em b}\kern-.08em
    T\kern-.1667em\lower.7ex\hbox{E}\kern-.125emX}}
\begin{document}
\makeatletter
\renewcommand{\@makefntext}[1]{%
  \makebox[0em][r]{\@thefnmark\ }
  #1
}
\makeatother
\renewcommand{\footnoterule}{
  \kern -3pt
  \hrule width 2in
  \kern 2.6pt
}
\renewcommand{\thefootnote}{}

\title{
Baichuan2-Sum: Instruction Finetune Baichuan2-7B Model for Dialogue Summarization 
}

\author{\IEEEauthorblockN{1\textsuperscript{st} Jianfei Xiao\textsuperscript{*}}
\IEEEauthorblockA{
\textit{Viterbi School of Engineering}\\
\textit{University of Southern California}\\
Los Angeles, United States \\
jianfeix@usc.edu}
\and
\IEEEauthorblockN{2\textsuperscript{nd} Yancan Chen\textsuperscript{*}}
\IEEEauthorblockA{
\textit{School of Computing} \\
\textit{National University of Singapore}\\
Singapore, Singapore \\
yancan@u.nus.edu}
\and
\IEEEauthorblockN{3\textsuperscript{rd} Yimin Ou}
\IEEEauthorblockA{\textit{College of Computing and Information Science} \\
\textit{Cornell University}\\
New York, United States. \\
yo243@cornell.edu}
\and
\IEEEauthorblockN{4\textsuperscript{th} Hanyi Yu}
\IEEEauthorblockA{\textit{Viterbi School of Engineering} \\
\textit{University of Southern California}\\
Los Angeles, United States \\
hanyiyu@usc.edu}
\and
\IEEEauthorblockN{5\textsuperscript{th} Kai Shu}
\IEEEauthorblockA{\textit{Viterbi School of Engineering} \\
\textit{University of Southern California}\\
Los Angeles, United States \\
kaishu.cs@gmail.com}
\and
\IEEEauthorblockN{6\textsuperscript{th} Yiyong Xiao\textsuperscript{\dag}}
\IEEEauthorblockA{\textit{Computer Science College} \\
\textit{Taiyuan Normal University}\\
Guangzhou, China \\
youchenyue@gmail.com }
}
\maketitle

\begin{abstract}
Large language models (LLMs) like LLaMA, Baichuan and Bloom models show remarkable ability with instruction fine-tuning in many natural language tasks. Nevertheless, for the dialogue summarization task, which aims to generate summaries for different roles in dialogue, most of the state-of-the-art methods foucus on small models (e.g BART and BERT). Existing methods try to add task specified optimization on small models like adding global-local centrality score to models. In this paper, we propose an instruction fine-tuning model: Baichuan2-Sum, for role-oriented dialogue summarization. By setting different instructions for different roles, the model can learn from the dialogue interactions and output the desired summaries. Furthermore, we applied NEFTune technique to add suitable noise during training, improving the results. The experiments demonstrate that the proposed model achieves the new state-of-the-art results on two public dialogue summarization datasets: CSDS and SAMSUM. The Baichuan2-Sum model shows an improvement in Rouge scores on both datasets compared to the previously best-performing model. Notably, for the SAMSUM dataset, there is a 21\% increase in the ROUGE-1 score, a 32\% increase in the ROUGE-2 score, and a 9\% increase in the ROUGE-L score. We have released our model and related codes to facilitate future studies in the dialogue summarization task.
\end{abstract}


\section {Introduction}
Dialogue summarization, a natural language processing (NLP) task, aims to generate a coherent summary from the content. With the development of the internet, the variety of dialogues (including human-to-human and human-computer conversations) and their volume have increased rapidly. It is valuable to generate the summarization automatically for dialogues like online customer service \cite{1,2} and meeting conversation \cite{3,4}. The figure 1 shows an example of a customer and service dialogue and its corresponding summarization from CSDS \cite{2} dataset.
\footnotetext{
* These authors contributed equally.\\ 
\dag \, Corresponding Author. \\
}
\begin{figure}[htbp]
    \centering
    \includegraphics[width=1\linewidth]{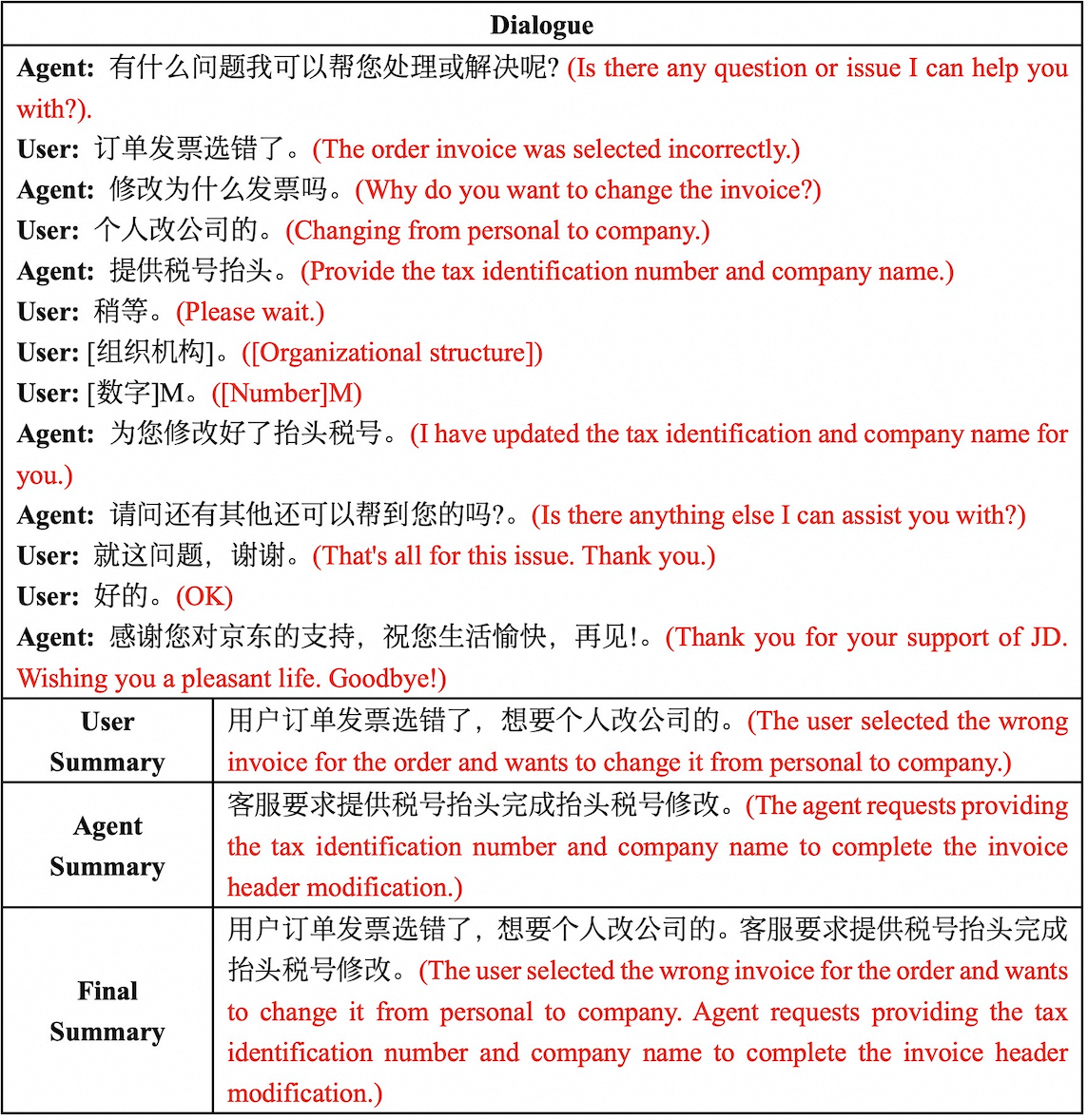}
    \caption{A customer service dialogue example of CSDS dataset}
    \label{fig1}
    \vspace{-15pt}
\end{figure}
\footnotetext{
Code address: https://github.com/aivolcano/LLM4DialogueSummarization
}
The generation of dialogue summaries presents many challenges, such as interactions between multiple roles\cite{5,6}, topic shifting between different contexts\cite{1,7,8} and understanding long dialogue context\cite{9}. To conquer these issues, the existing works focus on adding task specified optimization on language models. 
Liang et al.\cite{5} proposed a Role-Aware Centrality (RAC) model, which requires re-weighting the hidden states by calculating the role-aware centrality. Zou et al. \cite{1} present a two-stage dialogue summarization method with a saliency-aware neural topic model(SATM). Compared to the classical transformer achitecture, SATM adopts their proposed topic-informed attention mechanism. DIALOGLM \cite{9} is a pre-trained Model designed for long dialogue understanding and summarization. But the model is a pre-trained via a window-based denosing task, which means it cannot be directly fine-tuned on open-source NLP model weights. Besides, the backbone networks for majority of these works are small models like PGN \cite{10}, BERT \cite{11} and BART \cite{12} rather than large language models (LLMs) e.g. GPT-3 \cite{13}, LLaMA2 \cite{14}, Baichuan2 \cite{15}, Glm \cite{16}. The increase in parameters has brought significant improvements in the capabilities for language models, resulting in more human-like text generation for a range of text generation tasks, eg. machine translation, text summarization and question answering. Hence, it is valuable to investigate LLMs that can more effectively understand and summarize dialogues.

The contribution of this paper can be summarized as follows: 1) we present BaichuanSum, a model based on Baichuan2, trained on dialogue dataset (CSDS \cite{2} and SAMSUM \cite{17}), which achieves the new start-of-the-art performance for the dialogue summarization task. 2). we create an instruction fine-tuning dataset based on the original datasets, containing different instructions for various summarization types.
3) we employ the Noisy Embedding Instruction Fine Tuning (NEFT) \cite{18} method to train the model, further improving the model performance for this task.
4) Our work is publicly avalible on GitHub, offering good extensibility. In addition to Baichuan2, the code also supports training and evaluation for LLMs like LLaMA2 \cite{15}, Bloom \cite{19} and ChatGLM \cite{16}.

\section{Related Works}
\subsection{Dialogue Summarization Works}
PGN (Pointer-Generator Networks) \cite{10} is a sequence-to-sequence network for summarization, featuring an encoder with a bidirectional LSTM layer and a decoder with another bidirectional LSTM layer. Abigail et.al enhanced this model in two innovative ways. Firstly, They implemented a hybrid pointer-generator network, which enables the copying of words directly from the source text through a pointing operation.Secondly, they utilized coverage mechanism to track the information summarized, thus reducing repetition.

Liu et.al \cite{20} proposed a new framework based on BERT for both extractive (BERTExt) and abstractive (BERTAbs) summarization models. Since our task is the abstractive summarization, only abstractive model is used as a baseline model in the experiment part.

BART \cite{12} is a denoising autoencoder for sequence-to-sequence tasks, including machine translation and text summarization. It combines a bi-directional encoder, similar to BERT and an auto-regressive decoder, like GPT. Bart is trained by corrupting text through various noising operations such as Token Masking, Token Deletion and Text Infilling, etc., and is then required to reconstruct the original text.

BART-GLC (Global-Local Centrality) \cite{8} is a topic-aware BART model for dialogue summarization task. The model calculates the global and local centrality score of each utterance using Kmeans algorithm. GLC score (multiplication of global centrality score and local centrality score) is then used to reweight the hidden state of the BART. 

\subsection{Instruction Finetuning}
Instruction fine-tuning \cite{21,22,23} of large language models (LLMs) or large multimodal models (LMMs) has been proved to improve the zero-shot learning ability in many tasks. This technology enhances the ability of large language models to follow instructions. 

Wang et.al \cite{21} introduce a framework called SELF-INSTRUCT for improving the instruction-following capabilities of pretrained language models. The framework generates instructions, input, and output samples from a language model, filters invalid or similar ones, and uses them to finetune the original model. The researchers apply the method to the vanilla GPT3 model and demonstrate a 33\% improvement over the original model on a task named SUPER-NATURALINSTRUCTIONS.

After training on a 52K instruction-following dataset, Alpaca-7B \cite{22} has exhibited behaviors similar to those of OpenAI’s text-davinci-003.In this task, due to certain limitation,  we manually designed some instructions for different roles and datasets, instead of automatically generating instructions from LLMs.

The paper \cite{23} shows the first attempt to use language-only GPT-4 to generate multimodal language-image instruction-following data. Haotian et al. introduce LLaVA, a instruction finetuned large multimodal model that connects a vision encoder and an LLM for general-purpose visual and language understanding. The experiments show that LLaVA demonstrates impressive multimodal chat abilities and achieves a new state-of-the-art accuracy when fine-tuned on Science QA. 

The work above inspires us to employ fine-tuning for the task of dialogue summarization. We have created a fine-tuning dataset specifically for this task.

\subsection{Noisy Embedding with NEFTune}
Noisy Embedding technology aims to improve the model performance of pre-trained models on downstream tasks by adding noise to the paramesters of pre-trained model before fine-tuning. This method also applied on modern LLMs \cite{38}. When evaluated using AlpacaEval, the Llama2-7B model, after adding noise, showed a 34.9\% improvement on the Alpaca dataset compared to the model without noise! \cite{36,37} Additionally, this technology reduces overfitting effectively. In the case of longer outputs, the repetition rate does not raise significantly; instead, it provides more details, thereby enhancing the quality of the output text. Thanks to its simplicity and effectiveness in enhancing model performance, Hugging Face website has integrated this technology into Transformer Reinforcement Learning (TRL) library \cite{39}. 

\section{Methodology}
Figure 2 shows the main structure of our proposed Baichuan2-Sum model, which consists of Noisy Embedding, Decoder-based Transformers with some modifications, including xFormers, RMS Norm, and Rotary Positional Embedding (RoPE). In this section, let us introduce them step by step. 

\subsection{Tokenizer}
Before further processing, Baichuan2 model requires Baichuan2 tokenizer to break down the text into tokens, positional ids and attention mask.The Baichuan2 model has faster inference speed and smaller memory cousumption compared to other open source LLMs due to it has highest compression rate (See Table 1). The compression rate of Baichuan2's tokenizer is  0.498, which is half that of LLaMA 2, resulting in Baichuan2's speed being nearly twice as fast.

The CSDS dataset is comprised ntirely of Chinese language corpora. Compared to other tokenizers, Baichuan2 offers two advantages due to its comprehensive vocabulary of Chinese words: 1) Under identical text conditions, the model processes fewer tokens, leading to faster fine-tuning and predition; 2) The shorter token lengths reduce the demands on GPU memory.

\begin{table}[htbp]
    \centering
    \caption{The vocab size and text compression rate of Baichuan2 tokenizer and other tokenizers}
    \begin{tabular}{c|c|c}
        \hline\hline
        Tokenizer & Vocab Size & Compression Rate\\
        \hline
        LLaMA 2 & 32,000 & 1.037 \\
        Bloom & 256,680 & 0.501 \\
        ChatGLM 2 & 64,794 & 0.527 \\
        Baichuan 1 & 64,000  & 0.570 \\
        Baichuan 2 & 125,696 & 0.498 \\
        \hline\hline
    \end{tabular}
    \label{tab:table1}
\end{table}
\subsection{NEFTune: Noisy Embedding Instruction Fine-tuning}
The tokens are converted into input embedding vectors, with some uniform noise added to prevent model overfitting and to achieve better model performance without incurring additional computational costs. NEFTune \cite{18} is an effective strategy for modeling. In every training step, NEFTune introduces noise to the input embedding vector. This noise is generated from a uniform function within the range of [-1, 1] and matches the shape of the embedding vector. The noise is scaled by a factor of \(\alpha / \sqrt{Ld}\), where \(\alpha\) is a hyper-parameter, and \(L\) represents the sequence length and \(d\) is the embedding dimension.

\begin{algorithm}
\caption{NEFTune: \textbf{N}oisy \textbf{E}mbedding \textbf{I}nstruction \textbf{F}inetuning}
\begin{algorithmic}
\State \textbf{Input:} $D = \{(x_i, y_i)\}_{i=1}^N$ tokenized dataset, embedding layer $emb(\cdot)$, rest of model $f_{\text{/}emb}(\cdot)$, model parameters $\theta$, loss($\cdot$), optimizer $opt(\cdot)$
\State NEFT Hyperparameter: base noise scale $\alpha \in \mathbb{R}^+$
\State Initialize $\theta$ from a pretrained model.
\Repeat\ $(X_i, Y_i) \sim D$ \hfill \Comment{sample a minibatch of data and labels }

    \State $X_{emb} \leftarrow emb(X_i), \mathbb{R}^{B \times L \times d}$ \hfill\Comment{batch size $B$, seq. length $L$, embedding dimension $d$}  
    \State $\varepsilon \sim \text{Uniform}(-1, 1), \mathbb{R}^{B \times L \times d}$ \hfill\Comment{sample a noise vector}
    \State $X'_{emb} \leftarrow X_{emb} + \left(\frac{\alpha}{\sqrt{Ld}}\right)\varepsilon$ \hfill\Comment{add scaled noise to embeds$^a$}
    \State $\hat{Y_i} \leftarrow f_{/emb}(X'_{emb})$ \hfill\Comment{make prediction at noised embeddings}
    \State $\theta \leftarrow opt(\theta, \text{loss}(\hat{Y_i}, Y_i))$ \hfill\Comment{train step, e.g., grad descent}
\Until stopping criteria met/max iterations.
\State \hrulefill
\State \footnotesize{$^a$If sequence lengths in a batch are not equivalent, then $L$ is a vector $\in \mathbb{Z}_{>0}^B$ and the scaling factor $\left(\frac{\alpha}{\sqrt{Ld}}\right)$ is computed independently for each sequence in batch.}
\end{algorithmic}
\end{algorithm}

\subsection{Baichuan2 Model With NEFTune}
After feeding the tokens into the model, the embedding is processed by the Decoder architecture. This includes RMS Normalization, and the use of an enhanced attention mechanism known as xFormers with Rotary Positional Embedding (RoPE). It also involves a residual network structure and the SwiGLU activation function. The Decoder processes the data in a loop which runs N times, after which the token probability distribution is obtained via a softmax function. Throughout this procedure, the NEFTune strategy is applied to fine-tune the model, culminating in the output of our proposed model, Baichuan2-Sum.
 
Figure 2 illustrates the framework of Baichuan2 model integrated with NEFTune. Similar to GPT series models and LLaMA2, Baichuan2 consists exclusivly of decoder-only Transformers. The proposed architecture can be mainly divided into three parts: the Embedding part, the Transformer Block part, and the Linear part. Although the composition is similar to the classical large language models, the proposed model still exhibits some distinct differences. 

\begin{figure}[htbp]
    \centering
    \includegraphics[width=1\linewidth]{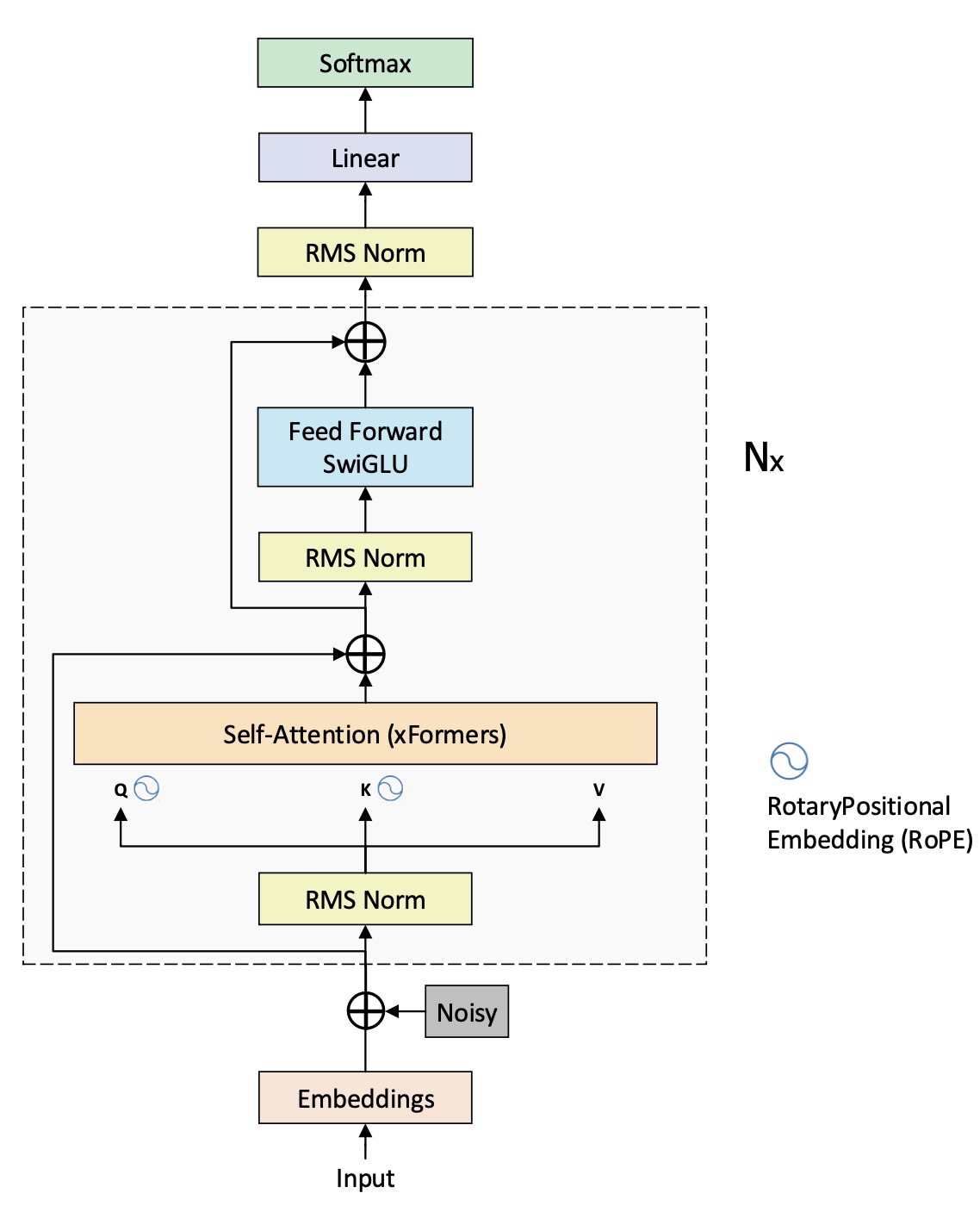}
    \caption{Baichuan2 Achitecture with NEFTuning}
    \label{fig:fig3}
\end{figure}

\textbf{Noisy Embedding:} We enhance the model's performance by introducing normal distribution noise into embedding using NEFTune method. The sole tunable parameter in NEFTune is \(\alpha\), we set it to 5 after numerous experiments. This setting demonstrated the best performance compared to other values.

\textbf{Tokenization:} Baichuan2 tokenizer has employed  SentencePiece \cite{34} and Byte-Pair Encoding (BPE) \cite{35} technologies. Unlike typical technology, it neither normalizes text nor adds a dummy prefix. Instead, it splits numerical values into single digits and adds a space token for code text with external spaces. Additionally, some infrequent characters have been converted back to the UTF-8 bytes. Thanks to these advanced technologies, the Baichuan2 tokenizer boasts a faster inference speed and lower memory consumption compared to other open-source LLMs.

\textbf{Rotary Positional Embedding (RoPE):} The Baichuan2-7B model adopts rotary positional embedding (RoPE) \cite{24} instead of typical positional embedding. This technique encodes absolute positions using a rotation matrix and explicitly integrates relative position information into the self-attention mechanism. RoPE is effective for handling long sequences as it can capture long-range dependencies more efficiently, thus enhancing the model performance with lengthy texts. Moreover, due to integrated with the self-attention mechanism, RoPE adds positional context without significantly increasing computational burden.

\textbf{SwiGLU:} Baichuan2 has employed SwiGLU \cite{25}, a switch-activated variant of Gated Linear Unit (GLU), to achieve the optimal value in log-perplexity, thereby  enhancing its generation results. 

\textbf{xFormer:} The authors of Baichuan 2 have adopted xFormers\cite{26}, which optimizes attention with biasing capabilities, to perform memory-efficient attention operations in the attention layers. This modification contributes to both performance enhancements and efficiency gains in the model’s large-scale training.

\textbf{RMSNorm:} RMSNorm \cite{27} is utilized to prevent the issues of vanishing or exploding gradients during training, leading to more stable training outcomes. Additionally, in comparison to Layer Normalization, RMSNorm typically requires less computational effort. This is becuase it computes only the root mean square of the inputs, rather than their mean, making the process more efficient, which lead to reducing computational time by 7\% to 64\% on different models. 

In table 3, we detail the parameters of the proposed model. The Baichuan2 model consists 32 transformer layers, each with 32 attention heads. It has a hidden size of 4,096 and a feed-forwarding dimension of 11,008. These dimensions are significantly larger than those of BART.
\begin{table}[htbp]
    \centering
    \caption{Details of Proposed Model}
    \begin{tabular}{c|c}
        \hline\hline
        Parameter & Value \\
        \hline
        Hidden Size & 4,096\\
        FFN Size & 11,008\\
        num heads & 32\\
        num layers & 32\\
        seq length & 4,096\\
        \hline
    \end{tabular}
    \label{tab:table2}
\end{table}
\section{Experiments}
\subsection{Dataset}
We evaluate our model on two dialogue summarization datasets: CSDS\textsuperscript{1} \cite{2} and SAMSUM\textsuperscript{2} \cite{17}. 

\textbf{CSDS} (\textbf{C}ustomer \textbf{S}ervice \textbf{D}ialogue \textbf{S}ummarization) is a Chinese dataset focused on role-based and real-world dialogue summarization. It not only produces overall dialogue summaries but also creates summaries specific to different roles within the dialogue. Specifically, the user summary represent the high-frequency questions, and the agent summary for the quality of customer service. The dataset includes 9,101 training samples and 800 samples each for testing and validation. 

\textbf{SAMSUM} (\textbf{S}amsung \textbf{A}bstractive \textbf{M}essenger \textbf{Sum}marization dataset) contains about 16,000 English dialogue summaries. These messenger-style conversations, crafted by linguists fluent in English, mirror the proportion of typical themes found in real-life messenger conversations. The dataset varies in style and register, ranging from informal and semi-formal to formal, and includes the use of slang, emoticons, and typographical errors. The summaries have aimed to encapsulate the discussions in a third-person narrative. The SAMSum includes 14,732 training samples, 818 validation samples, and 819 test samples.  
\begin{figure}[ht]
    \centering
    \includegraphics[width=1\linewidth]{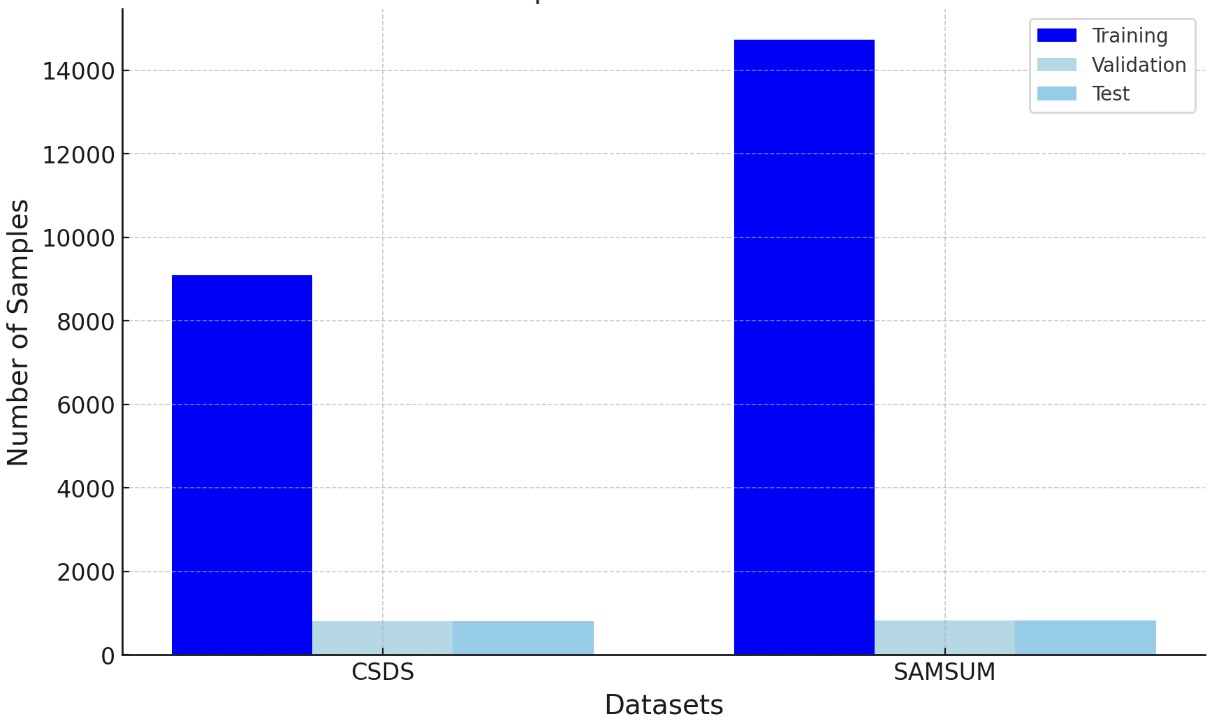}
    \caption{Distribution of Samples in CSDS and SAMSUM Datasets}
    \label{fig:fig4}
\end{figure}

\begin{figure}
    \centering
    \includegraphics[width=1\linewidth]{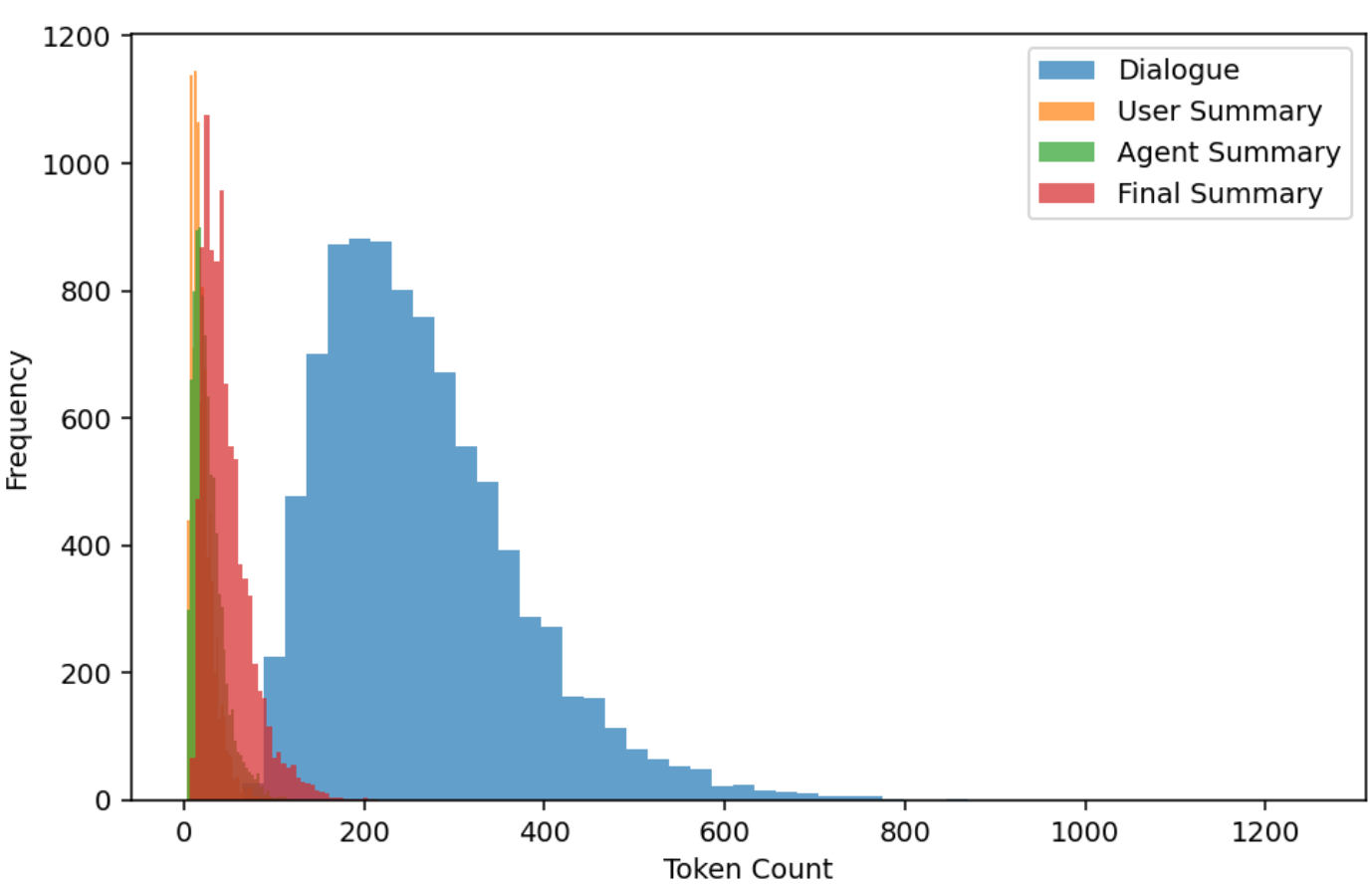}
    \caption{Token Length Distribution for CSDS Dataset}
    \label{fig:fig5}
\end{figure}

\begin{figure}
    \centering
    \includegraphics[width=1\linewidth]{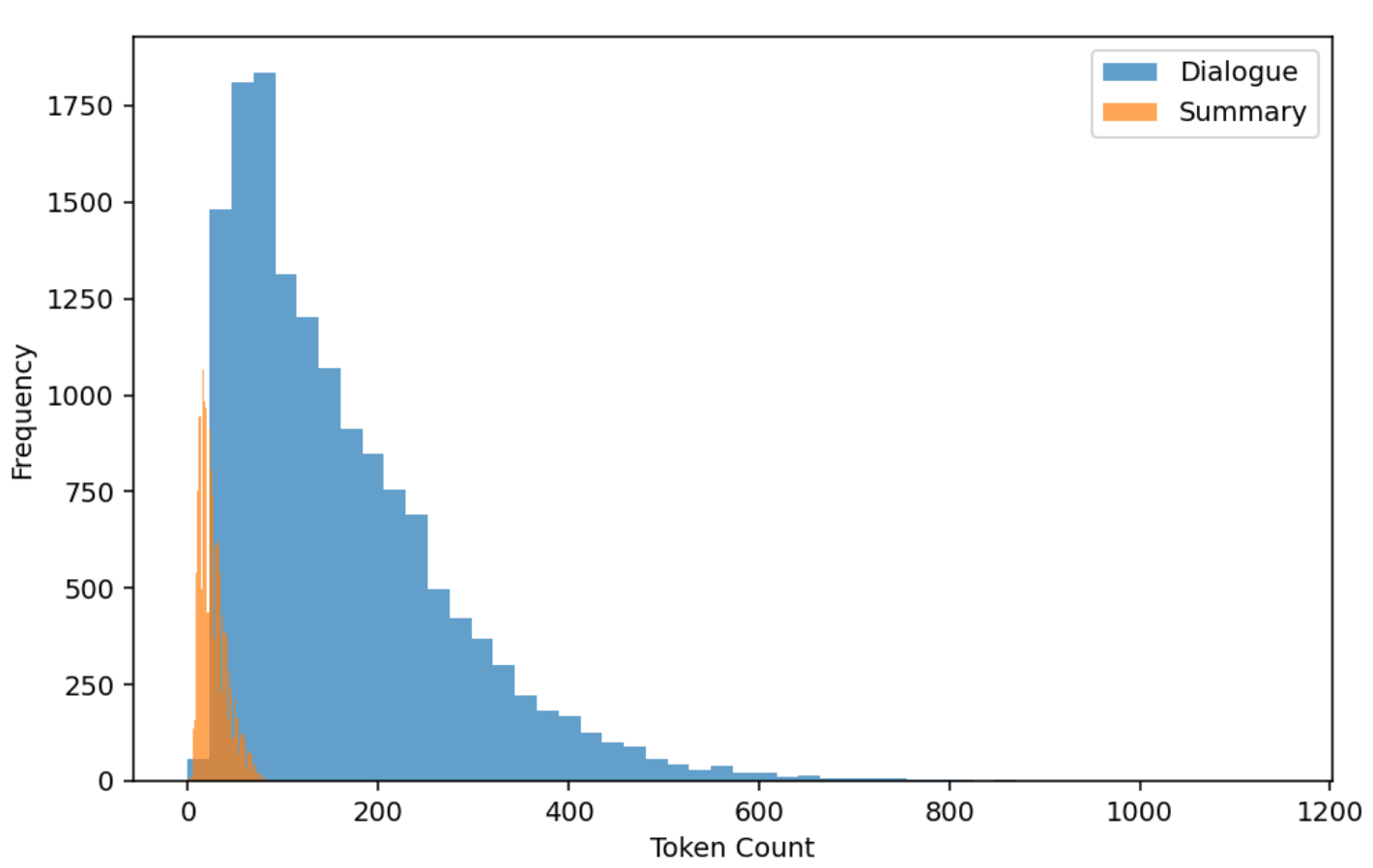}
    \caption{Token Length Distribution for SAMSUM Dataset}
    \label{fig:fig6}
\end{figure}

Figure 3 shows the training/validation/test sample distributions of both CSDS and SAMSUM datasets.
Meanwhile, in order to set the maximum token length for the Baichuan2 model, we have separately compiled the token length distributions for the dialogue and summary sections of the two datasets, as shown in Figures 5 and 6. From these figures, it is evident that the summary token length in both datasets are shorter than those of the dialogues, and the length of the dialogues is mostly less than 1200 tokens.
\footnotetext{
$^1$ https://github.com/xiaolinAndy/CSDS \\
$^2$ https://huggingface.co/datasets/samsum \\
}

\subsection{Instruction finetuning prompt}
Table 3 shows the instruction fine-tuning prompts for different datasets. Given that CSDS includes three types (Agent/User/All) of summarization, we have designed three distinct prompt templates. The instructions for CSDS are in Chinese, while those for SAMSUM are in English. This approach maintains the consistency of the language used in the instruction, dialogue, and summarization. We add the token ``\textless{}s\textgreater{}'' at the beginning of the template and the same token at the end to indicate the start and end of the generated sequence, respectively. These instruction templates style are designed to be straightforward and easily applicable to new dataset, demonstrating the transferability of this method.

\begin{CJK}{UTF8}{gbsn}
\begin{table*}
\begin{center}
\caption{Instruction finetuning Prompt}
\begin{tabular}{l|c|l}
\hline
Dataset & Type & Instruction Prompt Template \\
\hline
\multirow{3}{*}{CSDS} & Agent & \textless{}s\textgreater{}下面是一段电商公司的客服和用户之间的对话，请你给出客服的摘要。\textbackslash n \{\{Dialouge\}\} \textbackslash n \{\{Agent Summary\}\} \textless{}s\textgreater{} \\
\cline{2-3} & User & \textless{}s\textgreater{}下面是一段电商公司的客服和用户之间的对话，请你给出用户的摘要。\textbackslash n \{\{Dialouge\}\} \textbackslash n \{\{User Summary\}\} \textless{}s\textgreater{}  \\
\cline{2-3} & All & \textless{}s\textgreater{}下面是一段电商公司的客服和用户之间的对话，请你给出全部的摘要。\textbackslash n \{\{Dialouge\}\} \textbackslash n \{\{All Summary\}\} \textless{}s\textgreater{} \\
\hline
SAMSUM & All & \textless{}s\textgreater{}Please make the summarization of following dialogue. \textbackslash n \{\{Dialouge\}\} \textbackslash n \{\{All Summary\}\} \textless{}s\textgreater{}  \\
\hline
\end{tabular}
\end{center}
\label{tab:table3}
\end{table*}
\end{CJK}
\subsection{Evaluation Metrics}
To assess model performance, we utilize both lexical and semantic-level metrics. Lexical metrics, such as ROUGE-1/2/L \cite{28} and BLEU\cite{29}, quantify N-gram-based similarity between model-generated summaries and reference summaries, spanning 1-gram to 4-gram evaluations that cover word accuracy to phrase representation. For semantic similarity, BERTScore\cite{30} calculates cosine similarity between embeddings of reference tokens and candidate tokens, derived from pre-trained language models. These metrics are employed to gauge model performance. Displayed results are the averaged outcomes from multiple model checkpoints.
\subsection{Implementation Details}
The initial weights \textsuperscript{3} of the model were downloaded from Baichuan2's model page on Hugging Face website. Instead of training the full set of 7 billion parameters of the model, we train only a small ratio of parameters using Low-Rank Adaptation (LoRA) \cite{31} method. LoRA involves freezing the pre-trained model weights and adding trainable Low-Rank decomposition matrices to every layer of the Transformer architecture. This approach significantly reduces the number of trainable parameters for subsequent tasks. 

The LoRA training for Baichuan2 is implemented by PEFT\textsuperscript{4} (Parameter-Efficient Fine-Tuning) library. PEFT facilitates the effective adjustment of pre-trained language models (PLMs) for a range of downstream applications, without the necessity to fine-tune every parameter of the model. The latest version of the PEFT library includes the implementations of LoRA \cite{31}, Prefix Tuning \cite{32} and P-Tuning \cite{33}.  Our configuration for LoRA is as follows: Lora rank is 9, Lora target is W\_pack and Lora Dropout is 0.9.
\footnotetext{
$^3$ https://huggingface.co/baichuan-inc/Baichuan2-7B-Base \\
$^4$ https://github.com/huggingface/peft \\ 
$^5$ https://pypi.org/project/rouge-score/ \\
$^6$ https://github.com/mjpost/sacreBLEU \\ 
$^7$ https://github.com/Tiiiger/bert\_score \\
$^8$ https://github.com/xnliang98/bart-glc \\
}
We fine-tune our model on the processed CSDS dataset for 9000 iterations, using a learning rate of 5e-5 and a batch size of 2, with a gradient accumulation step of 4. The Adam optimizer was employed without weight decay, along with a cosine learning rate strategy that includes no warm-up steps. Gradient accumulation is a practical method to save GPU memory. It involves accumulating the gradients computed over multiple mini-batches before conducting a single parameter update. To prevent gradient explosion, gradient clipping by the L2 norm, as shown in the following formula, during training process. 
\[
   \mathbf{g'} = \frac{\mathbf{g}}{\max(1, \frac{\|\mathbf{g}\|}{\theta})} 
\]
where $\mathbf{g}$ is the gradient vector and $\theta$ is the threshold, the clipped gradient $\mathbf{g'}$, $\|\mathbf{g}\|$ represents the L2 norm of the gradient vector. 

The $\theta$ is set to 1 for clipping the L2 norm of gradient of whole model. To save memory and accelerate the training speed, we apply FP16 data type for all experiments. 
With the previous setting, we train Baichuan2-7B model with only one RTX 4090 GPU card. Instruction finetuning on CSDS dataset completes within 6 hours and on SAMSUM dataset completes within 4 hours.
To sum up, the specific details of experiments are shown in table 4.

\begin{table}[htbp]
    \begin{center}
    \caption{Details of experiments}
    \begin{tabular}{c|c}
        \hline\hline
        Parameter & Value \\
        \hline
        Learning Rate & 5e-5\\
        Iteration & 9000 \\
        Batch Size & 2\\
        Accumulation Step & 4\\
        Maximum L2 Norm & 1 \\
        Optimizer & Adam \\
        Learning Strategy & Cosine Learning \\
        Lora Rank & 9 \\
        Lora Target & W\_pack \\
        Lora Dropout & 0.1 \\
        \hline
    \end{tabular}
    \label{tab:table4}
    \end{center}
\end{table}
\subsection{Automatic Evaluation Results and Analysis}
We evaluates the all models introduced in the related work, including PGN, BERT, BART-GLC and our proposed model Baichuan2Sum. BART-GLC\textsuperscript{8}  is the previous state-of-the-art model for the dialogue summarization. Each block has three values, in a order of the following: final summary/user summary/agent summary. According to the results in Table 5, Baichuan2Sum outperforms the other models in all types of summary in terms of ROUGE-1/2/L and BERTScore metrics. Although the BLEU score of Baichuan2Sum for the final summary and agent summary is not as good as Bart-GLC, the gap between them is not very large. For the text summarization task, especially for the diaglogue summarization with multiple role, the ROUGE-1 scores of Baichuan2Sum (60.72/63.01/56.21) are high for practical application in real-world scenarios. 
\begin{table*}[htbp]
\begin{center}
\caption{Evaluation on CSDS Dataset}
\begin{tabular}{l|c|c|c|c|c}
\hline\hline
CSDS & ROUGE-1 & ROUGE-2 & ROUGE-L & BLEU & BERTScore \\
\hline
PGN & 55.58/53.55/50.20 & 39.19/37.06/35.12 & 53.46/51.05/47.59 & 30.03/29.64/28.25 & 77.96/ 78.68/76.13 \\
BERT & 53.87/52.72/49.57 & 37.59/36.39/33.82 & 52.40/50.44/46.83 & 29.90/30.17/26.99 & 78.52/ 79.23/76.39 \\
BART & 59.07/58.78/53.89 & 43.72/43.59/40.24 & 57.11/56.86/50.85 & 34.33/34.26/31.88 & 79.74/ 80.67/77.31 \\
BART-GLC & 60.07/61.42/54.59 & 44.67/45.83/40.02 & 58.10/59.25/52.43 & 35.89/36.43/32.58 & 80.10/ 81.83/77.61 \\
\textbf{Baichuan2-Sum} & \textbf{60.36/62.97/56.08} & \textbf{45.33/47.05/41.11} & \textbf{58.69/60.67/53.55} & 33.90/\textbf{36.92}/30.79 & \textbf{80.53/82.10/77.61} \\
\textbf{Baichuan2-Sum(NEFTune)} & \textbf{60.72/63.01/56.21} & \textbf{45.50/47.53/41.36} & \textbf{58.66/60.92/53.83} & 33.92/\textbf{36.84}/30.96 & \textbf{80.67/82.48/78.68} \\
\hline\hline
\end{tabular}
\end{center}
\label{tab:table5}
\end{table*}

The Table 6 shows the evaluation results of both baseline models and Baichuan2Sum model on SAMSUM dataset. Compared to CSDS dataset, SAMSUM dataset is less complex, as it has a shorter average token length and only one type of summarization, resulting in relatively higher scores over all metrics. The Rouge-1, Rouge-2, and Rouge-L scores of Baichuan2 model have seen significant improvement compared to other models, such as Bart-GLC. This indicates that the generated text has a high degree of lexical similarity with the reference text. Compared to other baseline models, Baichuan2Sum's BLEU and BERT Score are at a moderate level despite significantly leading in Rouge scores. This could be due to a lack of diversity in the generated text. 
\begin{table*}[htbp]
\begin{center}
\caption{Evaluation on SAMSUM Dataset}
\begin{tabular}{l|c|c|c|c|c}
\hline\hline
SAMSUM & ROUGE-1 & ROUGE-2 & ROUGE-L & BLEU & BERTScore \\
\hline
PGN & 40.08 & 15.28 & 36.63 & 37.49 & 80.67 \\
BERT & 50.34 & 24.71 & 46.63 & 46.98 & 88.72 \\
BART & 53.12 & 27.95 & 49.15 & 49.28 & 92.14 \\
BART-GLC & 53.74 & 28.83 & 49.62 & 50.36 & 92.77 \\
\textbf{Baichuan2-Sum} & \textbf{74.38} & \textbf{60.72} & \textbf{58.21} & 46.74 & 83.28 \\
\textbf{Baichuan2-Sum(NEFTune)} & \textbf{74.51} & \textbf{60.87} & \textbf{58.26} & 46.51 & 84.19 \\
\hline\hline
\end{tabular}
\end{center}
\label{tab:table6}
\end{table*}

\subsection{Human Evaluation Results and Analysis}
We adopt the human evaluation ways proposed by Liang et. al \cite{8} to access the performance on the CSDS and SAMSUM datasets. Specifically, we randomly select 100 samples from each dataset and invite five NLP researchers to conduct a comparative analysis. The comparison is proposed model over previous best model Bart-GLC and the result will fall into three cases: win, tie and loss, meaning that our model is better, equal and weaker to the Bart-GLC model. The evaluation standards focus on the three aspects:
\begin{itemize}
    \item \textbf{Accuracy}: This refers to how well the summary captures the essential information and facts presented in the original text. 
    \item \textbf{Coherence}: A good summary should be coherent and logically structured. It should flow smoothly, with ideas and sentences connected in a way that makes sense.
    \item \textbf{Grammatical Correctness}: The summary should be free of grammatical errors, ensuring it is professionally presented.
\end{itemize}
We collect research results and draw the results in Figure 7. 
From the Figure 7, we can draw two conclusions. The Baichuan2Sum model demonstrates better performance than BART-GLC on both datasets. Compared to the CSDS dataset, the Baichuan2Sum model shows a greater lead in performance on the SAMSUM dataset.
\begin{figure}
    \centering
    \includegraphics[width=1\linewidth]{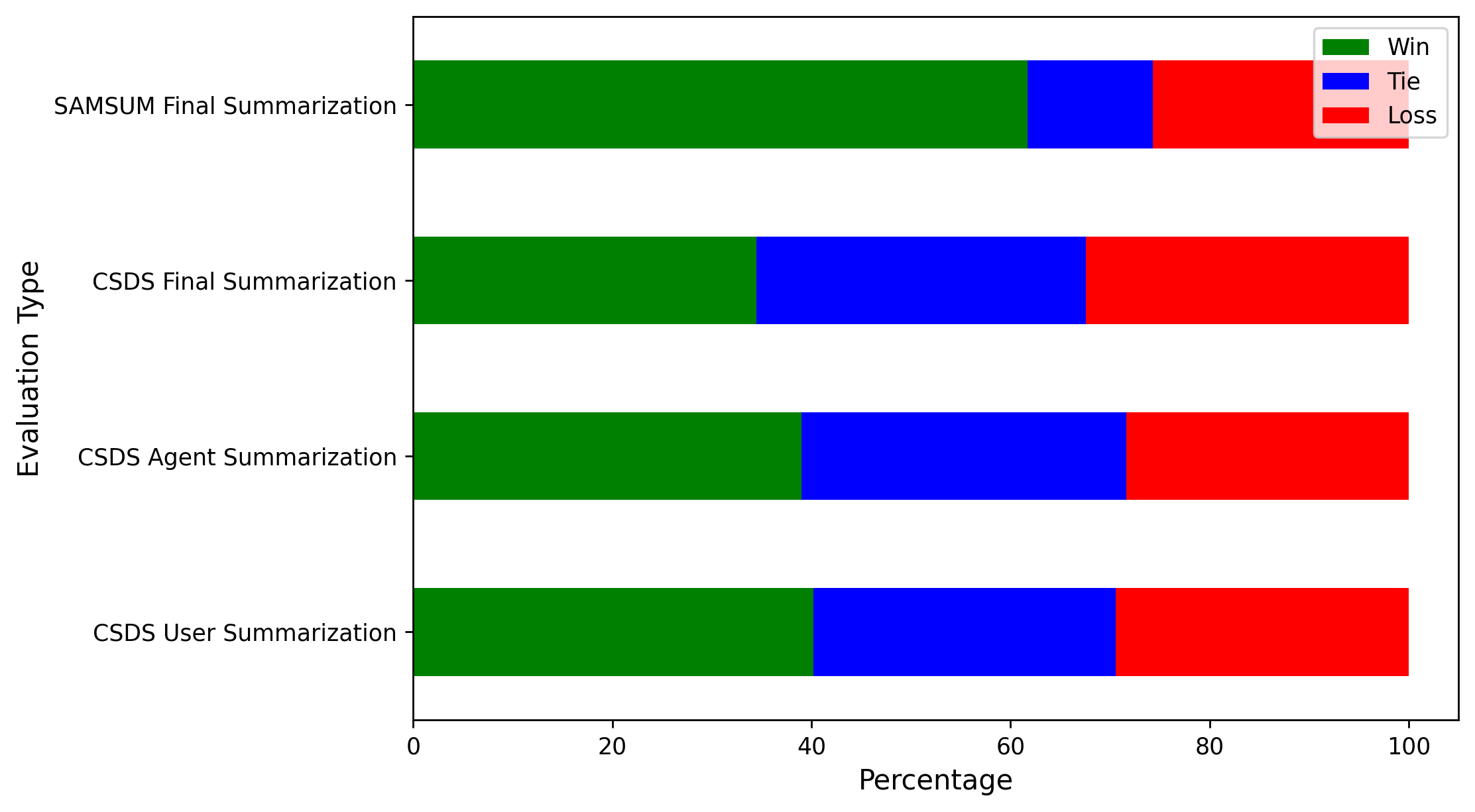}
    \caption{Human Evaluation Result on CSDS and SAMSUM dataset}
    \label{fig:fig7}
\end{figure}

\section{Limitation}
Even though our model achieves good result for the two dataset, there still exists some limitations for our work. 
Due to the limitation of computer resource, we only trained on the Baichuan2-7B model. Firstly, many recent NLP studies show that model size can affect the result and a larger model usually has a better performance. Hence, to improve the generation result, more large language models which have more parameters can be explored. Secondly, the experiment indicates that the proposed model still has room for improvement on both the BLEU and BERT Score metrics. Given that the diversity of text generation is positively correlated with these two metrics, we can use higher temperature scores and top-p values during the model's generation, or increases the beam size for beam search.

\section{Conclusion}
In this paper, we instruction finetune the Baichuan2-7B model for two dialogue summarization datasets. Before training model, we create various instruction templates and generated corresponding datasets. Besides, the NEFTune technique is applied to add noisy on embedding layer to improve the model performance. 

In the CSDS dataset, the Baichuan2Sum (NEFTune) model generally outperforms the other model across ROUGE-1, ROUGE-2, ROUGE-L, BLEU, and BERTScore metrics, showing an approximately 1\% increase in each. The model also demonstrates the superior performance in all metrics on SAMSUM dataset, particularly improvements of 21\% in ROUGE-1, 32\% in ROUGE-2, 9\% in ROUGE-L compared to other baselines. Additionally, human experiments show that our proposed model has best performance in summarization task across accuracy, coherence, and grammatical correctness metrics. 

The model is also effective, which is trained on only one RTX4090 GPU in less than 6 hours for CSDS and 4 hours for SAMSUM. Finally, to facilitate research by a wider audience, we have made our code publicly available on GitHub and provided details about the model and experiments in our paper.

\end{document}